\documentclass[conference]{IEEEtran}
\IEEEoverridecommandlockouts

\usepackage{cite}
\usepackage{hyperref}
\usepackage{amsmath,amssymb,amsfonts}
\usepackage{algorithmic}
\usepackage{graphicx}
\usepackage[a4paper, left=15mm, right=15mm, top=25mm, bottom=30mm]{geometry}  
\usepackage{textcomp}
\usepackage{xcolor}
\usepackage{booktabs} 
\usepackage[numbers,sort&compress]{natbib} 
\usepackage{cite} 
\usepackage{multirow} 
\usepackage{amsmath}
\usepackage{colortbl}
\usepackage{color}

\definecolor{LightGray}{gray}{0.93}
\definecolor{Green}{rgb}{0, 0.6, 0}
\definecolor{Red}{rgb}{0.85, 0.1, 0.1} 

\def\BibTeX{{\rm B\kern-.05em{\sc i\kern-.025em b}\kern-.08em
    T\kern-.1667em\lower.7ex\hbox{E}\kern-.125emX}}
\begin{document}

\title{M3D: Dual-Stream Selective State Spaces and Depth-Driven Framework for High-Fidelity Single-View 3D Reconstruction\\

}
\author{{\Large
Luoxi Zhang\textsuperscript{1,*}, Pragyan Shrestha\textsuperscript{1}, Yu Zhou\textsuperscript{3}, Chun Xie\textsuperscript{2}, Itaru Kitahara\textsuperscript{2}}
\\ 
\\ 
\textsuperscript{1}Doctoral Program in Empowerment Informatics, University of Tsukuba, Japan\\
\textsuperscript{2}Center for Computational Science, University of Tsukuba, Japan\\
\textsuperscript{3}Department of Data Science and Artificial Intelligence, The Hong Kong Polytechnic University, China\\
}

\maketitle

\begin{abstract}
The precise reconstruction of 3D objects from a single RGB image in complex scenes presents a critical challenge in virtual reality, autonomous driving, and robotics. Existing neural implicit 3D representation methods face significant difficulties in balancing the extraction of global and local features, particularly in diverse and complex environments, leading to insufficient reconstruction precision and quality. We propose M3D, a novel single-view 3D reconstruction framework, to tackle these challenges. This framework adopts a dual-stream feature extraction strategy based on Selective State Spaces to effectively balance the extraction of global and local features, thereby improving scene comprehension and representation precision. Additionally, a parallel branch extracts depth information, effectively integrating visual and geometric features to enhance reconstruction quality and preserve intricate details. Experimental results indicate that the fusion of multi-scale features with depth information via the dual-branch feature extraction significantly boosts geometric consistency and fidelity, achieving state-of-the-art reconstruction performance. Our code and dataset are publicly available at \href{https://github.com/AnnnnnieZhang/M3D}{\textit{this URL}}.

\end{abstract}

\begin{IEEEkeywords}
Single-View 3D Reconstruction, Implicit Neural Representation, Selective State Spaces Model, 3D Vision
\end{IEEEkeywords}

\section{Introduction}
\label{sec:intro}

The demand for high-fidelity 3D reconstruction from single-view images has surged in recent years, driven by its extensive applications in virtual reality, autonomous driving, and robotics \cite{VR_Applications, Autonomous_3D_Reconstruction}. Single-view 3D reconstruction aims to infer the complete 3D structure of objects from a single RGB image, a challenging task due to inherent ambiguities and the lack of depth information \cite{3D_Reconstruction_Review}. To achieve accurate and robust 3D reconstruction, models must effectively capture global and local features to represent object structures while preserving scene complexity \cite{Deep3D_SceneUnderstanding}.

Existing methods have shown progress in this area, yet significant limitations remain. Convolutional neural networks (CNNs), widely used in 3D reconstruction, excel at extracting local features. However, their limited receptive fields hinder their ability to capture global context, often resulting in incomplete or distorted geometry in complex scenes with occlusions \cite{CNN_LocalFeatures}. On the other hand, transformer-based architectures capture long-range dependencies effectively but often fail to model intricate local details, especially when reconstructing objects with complex geometries \cite{Transformers_GlobalFeatures}. These limitations highlight a primary bottleneck in improving the quality of single-view 3D reconstruction.

To address these issues, recent works have proposed various approaches focused on enhancing feature extraction. For instance, Li et al. \cite{MGN} introduced a joint layout and object mesh reconstruction method to achieve a comprehensive 3D understanding from single images. Similarly, Zhou et al. \cite{LIEN} employed implicit representations to model holistic scene understanding. InstPIFU \cite{InstPIFU} targets high-fidelity single-view reconstruction with pixel-aligned features, focusing on fine-grained details. SSR \cite{SSR} recently presented a high-fidelity 3D scene reconstruction framework that emphasizes capturing shape and texture from single-view images. While these methods have contributed significantly to advancements in 3D reconstruction, challenges in achieving balanced feature extraction for both global structure and local details remain. Introducing depth information helps address challenges such as resolving ambiguities in occluded regions and improving geometric consistency by enhancing geometric consistency and resolving ambiguities in complex scenes. Therefore, we designed a dual-stream structure to effectively incorporate depth information into the model.

Building on the above limitations of previous works \cite{SSR, InstPIFU}, we propose the M3D framework, a novel dual-streaM feature extraction framework for single-view 3D reconstruction. This novel approach addresses the challenges of single-view 3D reconstruction through a dual-stream feature extraction strategy. Our framework integrates a Selective State Space Model (SSM) \cite{SSM} that combines optimized residual convolutional layers \cite{residual_block} for capturing shallow features and a transformer-based component \cite{transformer_paper, mambavision_paper} for long-range contextual information. This design effectively balances the extraction of global and local features, enhancing the model's understanding of complex scenes. Observing the impact of occlusion on reconstruction accuracy, we further introduce a dedicated depth estimation stream to supplement RGB features with precise geometric context. This dual-stream structure enables the model to capture complementary information from RGB and depth data, leading to significant improvements in reconstruction quality and detail accuracy. Our depth estimation module utilizes a state-of-the-art single-image depth estimation model trained on large-scale data. This offline-generated depth map improves accuracy while reducing computational burden, providing a consistent geometric foundation for feature extraction. Our experiments demonstrate that M3D performs better on complex scenes with occlusions and intricate details, significantly surpassing existing methods. Specifically, M3D achieves a 36.9\% improvement in Chamfer Distance (CD) \cite{cd}, a 13.3\% increase in F-score, and a 5.5\% boost in Normal Consistency (NC) \cite{nc1,nc2} over the baseline methods on the Front3D dataset (Table~\ref{tab:3dfront}). Our main contributions are summarized as follows:
\begin{itemize}
    \item We propose a dual-stream feature extraction strategy that balances global and local information extraction, enhancing the quality of single-view 3D reconstruction.
    \item We design a depth estimation module, 'Depth Anything', which generates precise depth maps to supplement visual features, providing a consistent and enriched geometric context to address inaccuracies caused by variations in color and texture in traditional RGB methods.
    \item Through extensive experiments, we demonstrate that the M3D framework achieves leading performance on complex scenes, such as occlusions and fine details, with state-of-the-art(SOTA) quantitative metrics and visual fidelity.
\end{itemize}

\section{Related Work}
\label{sec:related_work}

\textbf{Single-View 3D Reconstruction Techniques.}
Single-view 3D reconstruction is a fundamental problem in computer vision, aiming to infer the complete 3D shape of an object from a single RGB image despite inherent ambiguity due to limited viewpoints. Early methods, such as voxel grids \cite{voxel1,voxel2} and graph cuts \cite{graph_cut_methods}, leveraged geometric priors to reconstruct simple structures, but faced limitations in complex scenes. Recent advances in neural implicit representations, such as Occupancy Networks \cite{occupancy_networks} and Neural Radiance Fields (NeRF) \cite{nerf}, have enabled continuous, high-quality reconstructions by implicitly modeling 3D shapes through neural networks. However, these methods struggle with complex backgrounds and occlusions, resulting in reduced accuracy and detail fidelity. To improve reconstruction, various NeRF variants, like Zero123 \cite{zero123} and Triplane-Gaussian \cite{triplane_gaussian}, were introduced to better capture geometric details. Despite these developments, challenges persist in balancing global and local feature extraction for fine-grained reconstructions in complicated environments. Based on these insights, we propose a selective SSM to achieve efficient global and local feature extraction, enhancing accuracy and fidelity in complex scenarios.

\noindent\textbf{Image Feature Extraction in 3D Reconstruction.}
Feature extraction is crucial in single-view 3D reconstruction, directly impacting the model's ability to capture both local details and global structure. Traditional methods relying on CNNs \cite{cnn}, including MGN \cite{MGN}, LIEN \cite{LIEN}, and InstPIFu \cite{InstPIFU}, effectively capture local details through convolutional layers. However, due to the limited receptive field of CNNs, these methods struggle to capture global context, resulting in incomplete reconstructions in occluded or complex scenes. Transformer architectures such as Vision Transformer (ViT)  \cite{transformer_paper}have been introduced to address this, providing long-range dependency modeling for improved structural understanding. Despite their ability to capture global context, Transformers are computationally expensive, with a quadratic complexity that limits scalability in high-resolution tasks.  Our proposed selective SSM model leverages a linear complexity self-attention mechanism, achieving efficient global feature extraction while maintaining the accuracy needed for detailed reconstructions.

\noindent\textbf{Depth-Guided Reconstruction Approaches.}
Depth information is essential in single-view 3D reconstruction, providing geometric cues that help mitigate the limitations of monocular vision. Early methods like PlaneRCNN \cite{planercnn} incorporated depth by detecting planar structures, which enhanced geometric consistency but struggled with non-planar objects. With the advancement of deep learning, depth estimation networks, such as MiDaS \cite{midas}, have been widely adopted to produce high-quality depth maps, providing critical geometric context. However, many methods extract RGB and depth features jointly within a single stream, which often results in feature interference, affecting spatial consistency. Recent developments have leveraged large-scale models to improve monocular depth estimation. Some methods are worth noting, such as Marigold \cite{marigold}, which repurposes diffusion-based image generators for depth estimation, and Depth Anything \cite{depthanything}, which utilizes extensive unlabeled data to improve depth accuracy across diverse scenes. Our dual-stream selective SSM framework independently processes RGB and depth data, ensuring high-fidelity reconstruction performance even in complex and occluded scenarios.

\begin{figure*}[t]
  \centering
  \includegraphics[width=0.9\linewidth]{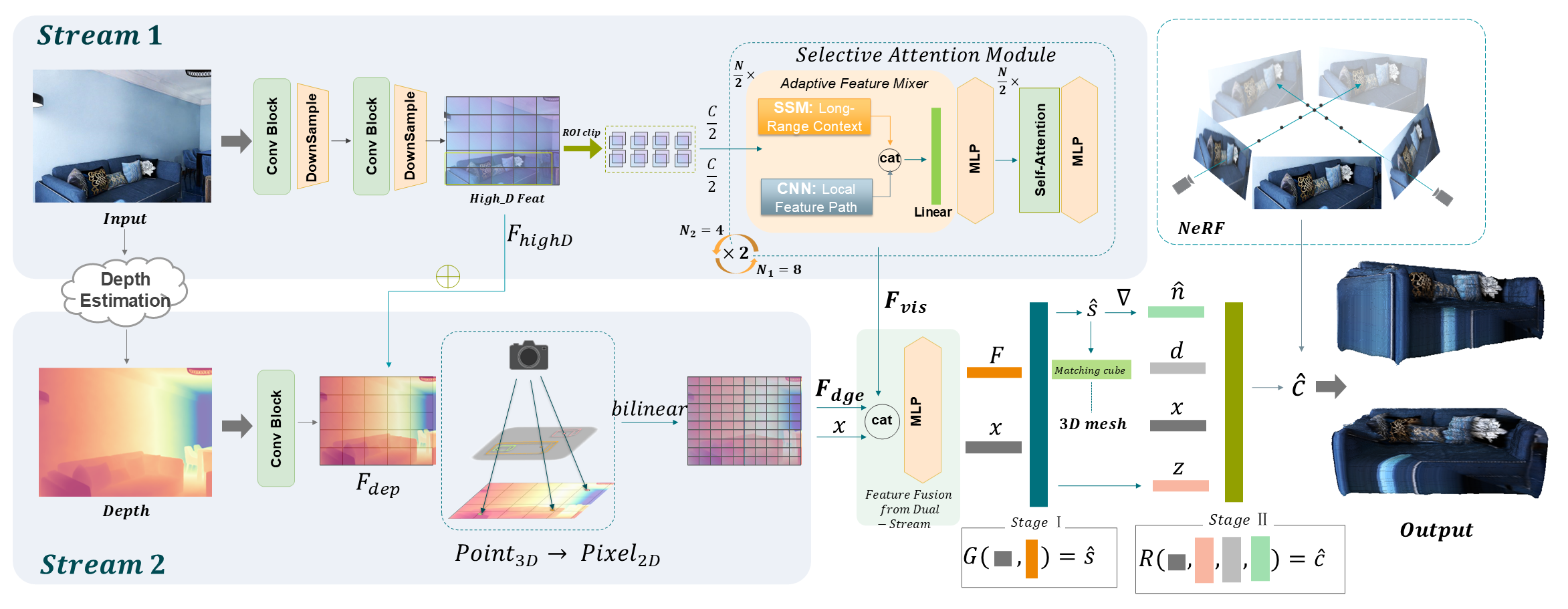}
  \caption{Overview of the M3D framework, showing the dual-stream architecture, selective attention module, implicit representation, and rendering network. This diagram highlights the core modules and the interaction between RGB and depth streams within the system.}
  \label{fig:method_flow}
\end{figure*}

\vspace{-0.5em}
\section{Method}
This section presents the M3D framework for high-fidelity 3D reconstruction from a single RGB image. The M3D framework integrates dual-stream feature extraction, a selective attention module \cite{mambavision_paper}, implicit geometric representation \cite{nerf}, and a rendering network to address the challenges of 3D reconstruction in complex geometries and occluded scenes.

\subsection{M3D Framework Overview}
The M3D framework comprises four main components: a dual-stream architecture, a Selective Attention Module, an implicit representation and geometric decoding module, and a loss supervision mechanism. First, the dual-stream architecture leverages an SSM for RGB feature extraction while incorporating a depth-driven stream to capture comprehensive geometric details. The Selective Attention Module integrates these long-range context and local features, producing spatially consistent representations that balance local details and global context. The implicit representation and geometric decoding module decodes high-dimensional features into 3D structures with high fidelity, leveraging volumetric rendering techniques. Finally, the loss supervision mechanism optimizes the framework, ensuring robust and accurate reconstructions by combining various supervisory signals. Each module contributes to the overall performance, enabling M3D to effectively handle complex scenes, producing accurate and detailed 3D models. Experimental results show that after introducing the depth-driven stream, the Chamfer Distance (CD) \cite{cd} improved by 16.3\% compared to the baseline. With the addition of the selective attention module, CD improved by 36.9\%, achieving state-of-the-art(SOTA) reconstruction performance.

\subsection{Dual-Stream Feature Extraction Module}
The dual-stream feature extraction module comprises an RGB feature extraction stream based on SSM and a depth-driven feature extraction stream. These two streams complement each other, providing enriched visual and geometric information for 3D reconstruction tasks.

\noindent\textbf{Selective Attention Module.} To capture both global and local information in the deep features, we introduce the Selective Attention Module. This module utilizes the SSM \cite{SSM} and a self-attention mechanism \cite{transformer_paper} to balance long-range dependencies and fine-grained details. Specifically, the high-dimensional ROI features \( F_{roi} \) are split into two parts: one half is processed by SSM to capture long-range dependencies, while a CNN processes the other half to extract local details \cite{mambavision_paper}. In the first iteration, the module has \( N_1 = 8 \) layers, \( N_1/2 = 4 \) layers dedicated to SSM and CNN in the Adaptive Feature Mixer and the remaining 4 layers to self-attention:
\begin{equation}
    F_{long} = \text{SSM}(F_{roi,1}), \quad F_{short} = \text{CNN}(F_{roi,2})
    \label{eq:ssm_cnn}
\end{equation}
where \( F_{roi,1} \) and \( F_{roi,2} \) are channel-split features from \( F_{roi} \). After concatenation and an MLP, the combined features are refined by self-attention:
\begin{equation}
    F_{context} = \text{SelfAttention}(\text{MLP}(F_{long} + F_{short}))
    \label{eq:self_attention}
\end{equation}
The second iteration uses \( N_2 = 4 \) layers. After processing in the first stage, the features already possess significant expressiveness, allowing further refinement with fewer layers without sacrificing performance. This iterative structure provides progressive feature refinement, allowing the module to revisit and enhance feature representations at multiple levels. By iterating between SSM and self-attention layers, the module achieves robust integration of global and local information, ensuring that long-range context and detailed characteristics are simultaneously captured. This design is particularly effective in complex scenes, where balancing global coherence with local detail is crucial.

\noindent\textbf{Depth-Driven Geometric Feature.}
We employ a depth-driven feature extraction stream to address the geometric ambiguity in single-view 3D reconstruction. Depth maps are generated offline using pre-trained models \cite{midas, marigold, depthanything}, which reduces the computational burden during training. The depth maps are then resized to match the high-dimensional feature input required for subsequent processing. To obtain robust geometric representations, we combine the depth-driven feature \cite{resnet} \( F_{dep} \) with the shallow feature \( F_{highD} \) from Stream 1 using a generalized addition operation, as shown in the figure. The combined feature is then projected onto the 2D plane based on the camera pose, with bilinear interpolation used to assign feature values:
\begin{equation}
  F_{3D} = \text{Bilinear}(F_{dep} \oplus F_{highD}, P_{2D})
  \label{eq:bilinear_mapping}
\end{equation}
where \( \oplus \) denotes the generalized addition operation between the depth and RGB-derived features, effectively bridging the RGB and depth modalities and enhancing the geometric feature representation. To fully exploit the complementary information from RGB and depth features, we concatenate the RGB-derived visual feature \( F_{vis} \), the depth-derived geometric feature \( F_{dge} \), and the interpolated 3D point coordinate \( x \). This 3D point \( x \) provides spatial context, allowing the RGB and depth features to align accurately with specific spatial locations in 3D space, thus enhancing the spatial consistency of the fusion. The concatenated feature is then passed through an MLP to learn the complex interrelations between these modalities:
\begin{equation}
  F_{fusion} = \text{MLP}(\text{concat}(F_{vis}, F_{dge}, x))
  \label{eq:fusion}
\end{equation}
This fusion strategy preserves local geometric details while enhancing global semantics, contributing to a more accurate and robust 3D reconstruction.

\noindent\subsection{Implicit Surface Modeling and Decoding}
\label{sec:implicit_geometry}

In this study, we encode 3D geometry using a neural implicit representation based on the Signed Distance Function (SDF) \cite{sdf}. For a given 3D point \( x \in \mathbb{R}^3 \), the SDF provides the distance \( s \) to the closest surface, where the sign indicates whether the point is inside (negative) or outside (positive) the surface. This can be expressed as follows:
\begin{equation}
    \text{SDF}(x) = s, \quad s \in \mathbb{R}
    \label{eq:sdf}
\end{equation}
The zero-level set of the SDF function, \(\Omega = \{x \in \mathbb{R}^3 | \text{SDF}(x) = 0\}\), implicitly represents the surface. The marching cube algorithm is used to transform every object's implicit SDF value into explicit meshes to generate the object meshes ~\cite{marchingcube}. To adapt the implicit representation for volumetric rendering, we convert the SDF value \( s \) to a density value \( \sigma \) to enable differentiable rendering \cite{sdfnerf}. A learnable parameter governs this conversion.This conversion is governed by a learnable parameter \( \beta \) and is defined as:
\begin{equation}
    \sigma_{\beta}(s) = 
    \begin{cases} 
        \frac{1}{2\beta} \exp\left(\frac{s}{\beta}\right), & s \leq 0 \\
        \frac{1}{\beta} \left(1 - \exp\left(-\frac{s}{\beta}\right)\right), & s > 0 
    \end{cases}
    \label{eq:density_conversion}
\end{equation}
For rendering, we sample \( M \) points along the ray \( r \) from the camera center \( o \) to the pixel in the viewing direction \( d \), where each sample point is defined as:
\begin{equation}
    x_i^r = o + t_i^r d, \quad i = 1, \dots, M
    \label{eq:sampling}
\end{equation}
with \( t_i^r \) representing the distance from the sample point to the camera center. After computing the SDF value \( s \) and color value \( c \) for each sample point, the predicted color \( \hat{C}(r) \) for ray \( r \) can be calculated as:
\begin{equation}
    \hat{C}(r) = \sum_{i=1}^{M} T_i \alpha_i c_i
    \label{eq:predicted_color}
\end{equation}
where \( T_i \) represents the transmittance and \( \alpha_i \) the alpha value at each sample point. These computations allow for efficient volumetric rendering on each ray, generating realistic color and geometric information.

\subsection{Training Supervision}
\label{sec:training_supervision}

We design a two-stage training supervision scheme that combines 3D geometry and 2D rendering supervision to improve model performance. In the first stage, only the 3D geometric supervision of the SDF is applied, with the loss function defined as:
\begin{equation}
    L_{3D} = \sum_{x \in X} |s(x) - \hat{s}(x)|
    \label{eq:3d_loss}
\end{equation}
where \( X \) is a set of uniformly sampled points along the rays and near the real surface. In the second stage, we introduce color loss, depth and normal consistency loss to enhance the supervision further. The photometric reconstruction loss \( L_{rgb} \) is defined as:
\begin{equation}
    L_{rgb} = \sum_{r} |C(r) - \hat{C}(r)|
    \label{eq:rgb_loss}
\end{equation}
Depth consistency loss \( L_{d} \) and normal consistency loss \( L_{n} \) leverage monocular geometric cues, assisting the model in reducing ambiguity in 3D shape reconstruction from single-view inputs. These losses are defined as:
\begin{equation}
    L_d = \sum_{r} ||D(r) - \hat{D}(r)||_2
    \label{eq:depth_loss}
\end{equation}

\begin{equation}
    L_n = \sum_{r} \left( ||N(r) - \hat{N}(r)||_1 + ||1 - N(r) \cdot \hat{N}(r)||_1 \right)
    \label{eq:normal_loss}
\end{equation}
The overall loss function combines these components, with weights \(\alpha\), \(\beta\), \(\gamma\), and \(\delta\) to control each term’s contribution:

\begin{equation}
    L = \alpha L_{3D} + \beta L_{rgb} + \gamma L_d + \delta L_n
    \label{eq:overall_loss}
\end{equation}
Here, \(\alpha\), \(\beta\), \(\gamma\), and \(\delta\) are hyperparameters. In our experiments, \(\alpha = 1\), while \(\beta\) and \(\gamma\) dynamically adjust to 0.1, and \(\delta\) adjusts to 0.01 during training. This supervision strategy allows the model to reconstruct high-quality 3D geometry and texture, balancing the focus on geometric and appearance details.

\section{Experiment}
\subsection{Experimental Setup} 
\textbf{Dataset.} We utilized the FRONT3D dataset \cite{front3d}, which offers various indoor 3D scenes suitable for single-view 3D reconstruction. The dataset comprises 29,659 images, and we adopt the same splits as Liu et al. \cite{split} for the datasets. 

\noindent\textbf{Training Settings.} All experiments were conducted on a single NVIDIA H100 GPU with a batch size of 30, utilizing PyTorch on Ubuntu 22.04. This setup maximized GPU memory efficiency and ensured consistent training stability. Key hyperparameters were optimized for model performance. The initial learning rate was set to 0.00006, tailored to the SSM framework \cite{SSM}, with a reduction to 0.00001 at epoch 110 for stable convergence by using StepLR. The model was trained for 200 epochs in total, and the best-performing model weights on the validation set were saved after each epoch to enhance generalization and mitigate overfitting.

\noindent\textbf{Evaluation Metrics.} To evaluate the model's performance on single-view 3D reconstruction, we employed several metrics: Chamfer Distance (CD) \cite{cd} for geometric fidelity, F-Score \cite{fscore} for reconstruction accuracy and completeness, Peak Signal-to-Noise Ratio (PSNR) \cite{PSNR} for visual quality, Intersection over Union (IoU) \cite{IoU} for spatial alignment, and Normal Consistency (NC) \cite{nc1,nc2}, for surface smoothness. These metrics together provide a comprehensive assessment across geometric precision, structural integrity, and visual fidelity.

\subsection{Experiment Results}

\begin{figure*}[t]
  \centering
  \includegraphics[width=0.84\linewidth]{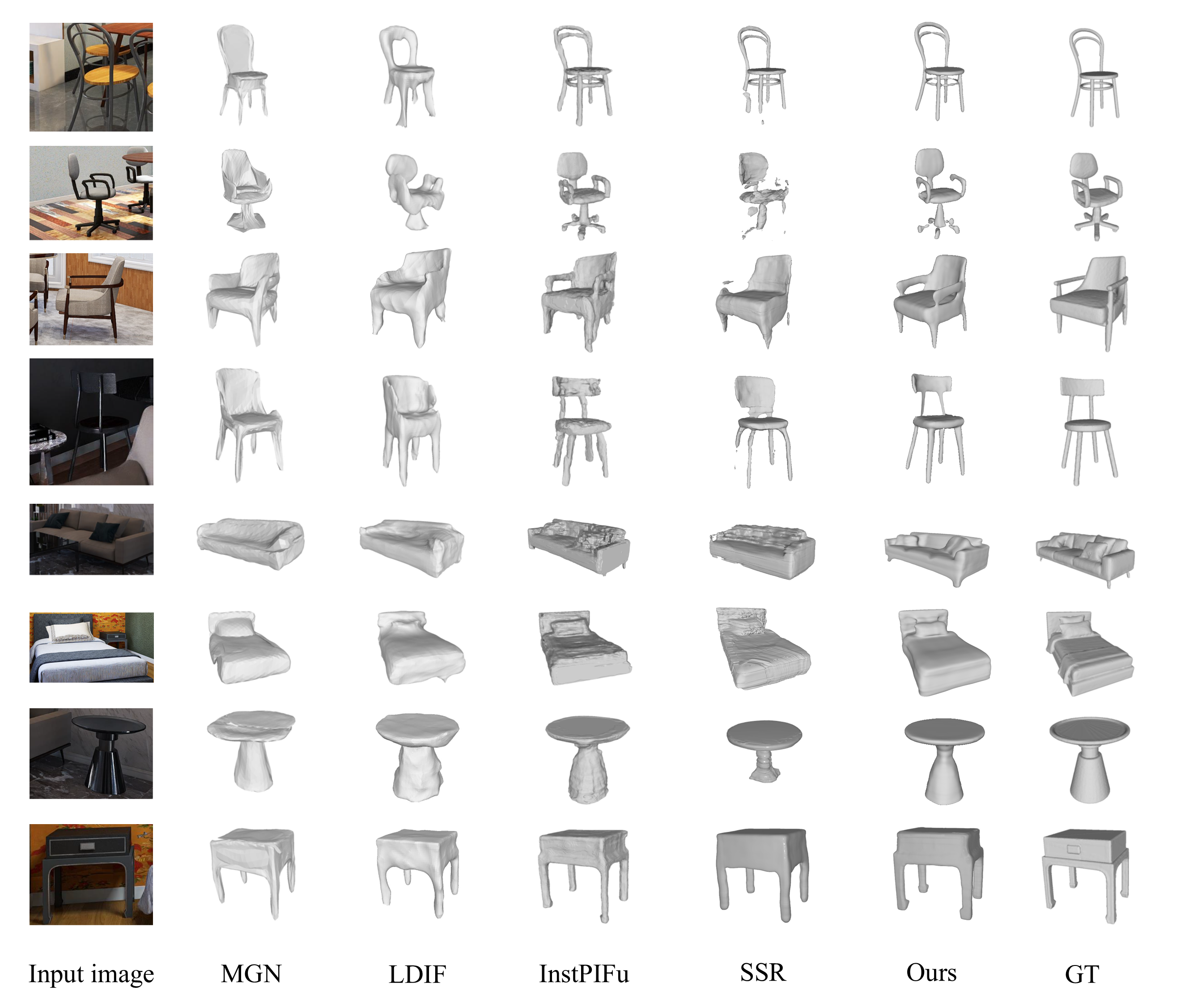}
  \caption{Comparison of 3D reconstruction results from different methods. Columns from left to right represent the input images, results from MGN\cite{MGN}, LDIF\cite{ldif}, InstPIFu\cite{InstPIFU}, SSR\cite{SSR}, our method, and the ground truth (GT). This figure demonstrates the improvement in geometric fidelity and completeness achieved by our method compared to the baseline approaches.}
  \label{fig:comparison_results}
\end{figure*}

\begin{table*}[t]
  \centering
  \caption{Evaluation of object reconstruction on the 3D-FRONT dataset~\cite{front3d}.}
  \label{tab:3dfront}

  \resizebox{0.85\textwidth}{!}{
    \begin{tabular}{@{}llccccccccc@{}}
      \toprule[1pt]
      \toprule[1pt]
      Metrics & Models & bed & chair & sofa & table & desk & nightstand & cabinet & bookshelf & mean \\
      \midrule[0.75pt]
      \multirow{2}{*}{\shortstack{CD $\downarrow$ \\ \textcolor{Red}{(51.0\%)}}} 
      & MGN \cite{MGN} & 15.48 & 11.67 & 8.72 & 20.90 & 17.59 & 17.11 & 13.13 & 10.21 & 14.07 \\
      & LIEN \cite{LIEN} & 16.81 & 41.40 & 9.51 & 35.65 & \underline{26.63} & 16.78 & 11.70 & 11.70 & 28.52 \\
      & InstPIFu \cite{InstPIFU} & 18.17 & 14.06 & 7.66 & 23.25 & 33.33 & \underline{11.73} & \underline{6.04} & 8.03 & 14.46 \\
      & SSR \cite{SSR} & \underline{4.96} & \underline{10.52} & \underline{4.53} & \underline{16.12} & \textbf{25.86} & 17.90 & 6.79 & \textbf{3.89} & \underline{10.45} \\
      \cmidrule(lr){2-11} 
      & \textbf{M3D \text{\small (ours)}} & \textbf{3.68} & \textbf{5.94} & \textbf{3.45} & \textbf{10.20} & 28.23 & \textbf{9.10} & \textbf{5.00} & \underline{5.45} & \textbf{6.60} \\
      \midrule[0.75pt]
      \multirow{2}{*}{\shortstack{F-Score $\uparrow$ \\ \textcolor{Red}{(33.6\%)}}} 
      & MGN \cite{MGN} & 46.81 & 57.49 & 64.61 & 49.80 & 46.82 & 47.91 & 54.18 & 54.55 & 55.64 \\
      & LIEN \cite{LIEN} & 44.28 & 31.61 & 61.40 & 43.22 & 37.04 & 50.76 & 69.21 & 55.33 & 45.63 \\
      & InstPIFu \cite{InstPIFU} & 47.85 & 59.08 & 67.60 & 56.43 & \underline{48.49} & 57.14 & \underline{73.32} & 66.13 & 61.32 \\
      & SSR \cite{SSR} & \underline{76.34} & \underline{69.17} & \underline{80.06} & \underline{67.29} & 47.12 & \underline{58.48} & 70.45 & \textbf{85.93} & \underline{71.36} \\
      \cmidrule(lr){2-11} 
      & \textbf{M3D \text{\small (ours)}} & \textbf{82.63} & \textbf{82.00} & \textbf{85.24} & \textbf{77.53} & \textbf{48.84} & \textbf{71.44} & \textbf{78.00} & \underline{81.66} & \textbf{80.85} \\
      \midrule[0.75pt]
      \multirow{2}{*}{\shortstack{NC $\uparrow$ \\ \textcolor{Red}{(10.3\%)}}} 
      & MGN \cite{MGN} & 0.829 & 0.758 & 0.819 & 0.785 & 0.711 & 0.833 & 0.802 & 0.719 & 0.787 \\
      & LIEN \cite{LIEN} & 0.822 & 0.793 & 0.803 & 0.755 & 0.701 & 0.814 & 0.801 & 0.747 & 0.786 \\
      & InstPIFu \cite{InstPIFU} & 0.799 & 0.782 & 0.846 & 0.804 & 0.708 & 0.844 & 0.841 & 0.790 & 0.810 \\
      & SSR \cite{SSR} & \underline{0.896} & \underline{0.833} & \underline{0.894} & \underline{0.838} & \underline{0.764} & \underline{0.897} & \underline{0.856} & \textbf{0.862} & 0.854 \\
      \cmidrule(lr){2-11} 
      &\textbf{ M3D \text{\small (ours)}} & \textbf{0.933} & \textbf{0.900} & \textbf{0.921} & \textbf{0.886} & \textbf{0.780} & \textbf{0.936} & \textbf{0.878} & \underline{0.842} & \textbf{0.901} \\
      \bottomrule[1pt]
      \bottomrule[1pt]
    \end{tabular}
  }
\end{table*}

\noindent\textbf{Geometric and Quantitative Analysis.}  
According 
to the quantitative results (Table~\ref{tab:3dfront}) and qualitative visual comparisons (Figure~\ref{fig:comparison_results}) reveals that M3D surpasses other methods~\cite{MGN,ldif,InstPIFU,SSR}. Compared to these methods, M3D achieves superior 3D reconstruction quality, showcasing enhanced smoothness, completeness, and detail accuracy that approach the fidelity of the GT. The metrics also exhibit strong improvements across key evaluation criteria.
 
The integration of the Selective SSM ~\cite{SSM} in M3D plays a crucial role in its superior performance. The SSM’s selective scanning operation architecture allows the model to dynamically adjust to input features by filtering irrelevant information and capturing essential contextual dependencies. This selective mechanism enhances both spatial consistency and robustness against noise, leading to smoother surface reconstructions and a 12\% improvement in NC values. Consequently, our model generates reconstructions with fewer artifacts and greater visual coherence.

The dual-stream architecture, incorporating both RGB and depth information, provides comprehensive geometric understanding by capturing complementary feature sets. This dual representation enables M3D to achieve an average 35\% increase in F-Score, highlighting its ability to reconstruct intricate details and maintain structural integrity. This feature fusion approach directly translates to improved reconstruction completeness and a higher level of surface continuity in the visual results.

blocks~\cite{mambavision_paper,residual_block} for shallow feature extraction contribute to a 56\% reduction in CD, indicating superior capture of geometric details. These optimized residual blocks effectively emphasize fine details and edges, leading to reconstructions that exhibit sharper and more accurate shapes than other methods. This improvement is evident in the visual results, where our reconstructions demonstrate a high degree of geometric fidelity and detail preservation. Overall, the results affirm that M3D achieves SOTA performance on the 3D-FRONT dataset ~\cite{front3d}, with each module contributing incrementally to superior quality in terms of reconstruction accuracy, structural completeness, and surface smoothness.

\begin{figure*}[h]
    \centering
    \includegraphics[width=0.82\linewidth]{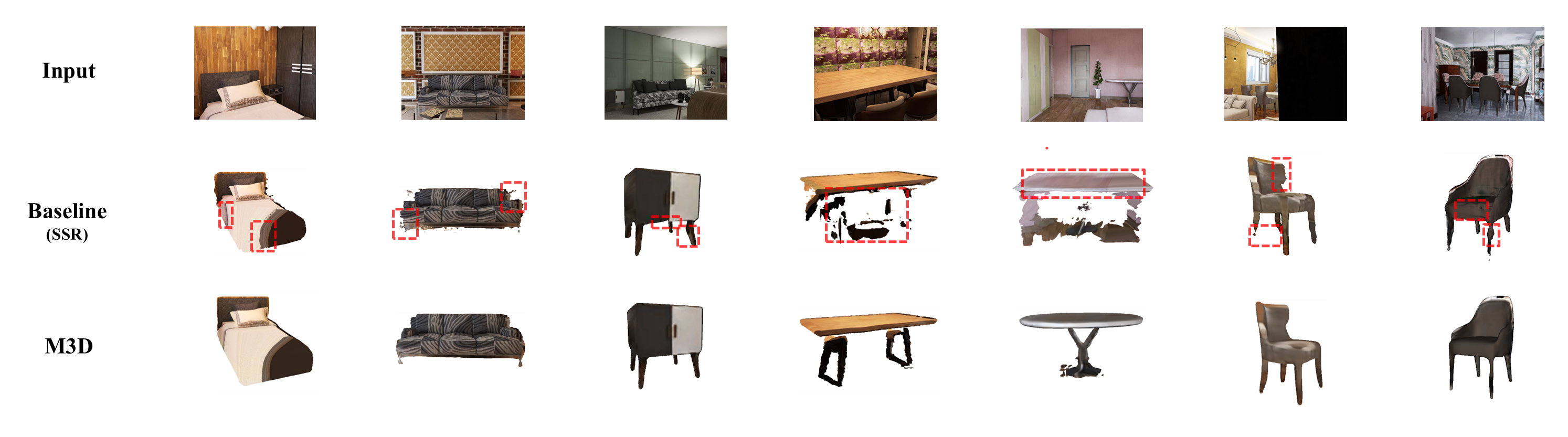}
    \caption{Texture comparison between M3D and SSR~\cite{SSR}, highlighting M3D's superior texture smoothness and color accuracy.}
    \label{fig:texture_comparison}
\end{figure*}

\noindent\textbf{Texture Analysis.} 
As shown in Figure \ref{fig:texture_comparison}, we selected SSR as the primary baseline due to its performance ranking second in metrics-based evaluations. While metrics effectively quantify geometric accuracy, they do not fully capture texture quality. Thus, we extended the comparison to assess texture continuity and color fidelity. The results demonstrate that our method surpasses SSR significantly in terms of texture smoothness and color perception. Additionally, our model’s superior geometric reconstruction, characterized by smooth and well-aligned surfaces, contributes to an enhanced texture reconstruction, yielding more realistic and visually coherent results.

\begin{table}[t]
  \centering
  \caption{Quantitative Comparison of Our Model with Zero-1-to-3 \cite{zero123} and Shape-E \cite{shape-e} on CD, F-Score, and NC Metrics.}
  \label{tab:comparison_metrics}
  \resizebox{0.7\columnwidth}{!}{ 
    \begin{tabular}{@{}lccc@{}}
        \toprule[1pt]
        \toprule[1pt]
      Methods & CD & F-Score & NC \\
      \midrule
      Zero-1-to-3 \cite{zero123} & 39.27 & 30.07 & 0.624 \\
      Shape-E \cite{shape-e}     & \underline{29.16} & \underline{39.86} & \underline{0.686} \\
      \textbf{M3D \text{\small (ours)}}       & \textbf{6.60} & \textbf{80.85} & \textbf{0.901} \\
      \bottomrule
    \end{tabular}
  }
\end{table}

\begin{table}[t]
  \centering
  \caption{Comparison of our method with recent 3D reconstruction methods on CD, F-Score, PSNR, and IoU metrics.}
  \label{tab:comparison}
  \resizebox{\columnwidth}{!}{ 
    \begin{tabular}{@{}lcccc@{}}
      \toprule[1pt]
      \toprule[1pt]
      Method & CD $\downarrow$ & F-Score $\uparrow$ & PSNR $\uparrow$ & IoU $\uparrow$ \\
      \midrule
      OpenLRM \cite{OpenLRM} & 0.0336 & 0.5354 & 18.0433 & 0.3947 \\
      InstantMesh \cite{InstantMesh} & 0.0161 & 0.6491 & 18.8262 & 0.5083 \\
      CRM \cite{CRM} & \underline{0.0141} & 0.6574 & 18.4407 & \underline{0.5218} \\
      Unique3D \cite{Unique3d} & 0.0143 & \underline{0.6696} & \underline{20.0611} & 0.5416 \\
      Unique3D w/o ET \cite{Unique3d} & 0.0158 & 0.6594 & 20.0383 & 0.5320 \\
      Wonder3D+ISOMER \cite{Wonder3D} & 0.0244 & 0.6088 & 18.6131 & 0.4743 \\
      \cmidrule(lr){1-5} 
      \textbf{M3D \text{\small (ours)} } & \textbf{0.0066} & \textbf{0.8085} & \textbf{30.0411} & \textbf{0.6128} \\
    \bottomrule[1pt]
    \bottomrule[1pt]
    \end{tabular}
  }
\end{table}

\noindent\textbf{Comparison with Advanced Single-View 3D Reconstruction Methods.} 
In the Table~\ref{tab:comparison_metrics}, we selected Zero-1-to-3~\cite{zero123} and Shape-E~\cite{shape-e} as baselines due to their utilization of large-scale pre-trained priors, which similarly leverage prior knowledge for 3D reconstruction. As shown in Table~\ref{tab:comparison_metrics}, our M3D method surpasses both Zero-1-to-3 and Shape-E across the CD, F-Score, and NC metrics, demonstrating clear advantages in reconstruction fidelity. This result suggests that, among methods utilizing pre-trained priors, our M3D framework—compared to Zero-1-to-3’s diffusion-based 2D priors and Shape-E’s textured mesh generation—offers a more effective approach for single-image 3D reconstruction, accurately capturing shape and texture details with higher precision and consistency.

Table \ref{tab:comparison} presents a comparison with recent popular 3D reconstruction methods \cite{OpenLRM,InstantMesh,CRM,Unique3d,Wonder3D}. Our model demonstrates clear superiority across these evaluation metrics \cite{cd,fscore,PSNR,IoU}. It highlights our model's advantages in both geometric accuracy, establishing it as a SOTA solution for 3D reconstruction tasks.

\section{Ablation}
We conduct four ablation experiments to assess the impact of feature extraction methods and dual-stream architecture on the final performance of our 3D reconstruction system.

\subsection{Ablation Experimental Setup}

\textbf{Baseline Experiment:} \\
The baseline experiment follows the setup of Chen et al. \cite{SSR}, using ResNet34 \cite{resnet} for shallow feature extraction and ResNet18 \cite{resnet} for deep feature extraction, with a single-stream RGB channel and no depth information.

\noindent\textbf{Experiment 1: Enhanced Residual Convolutional Blocks for Shallow Features} \\
To improve shallow feature extraction, we replaced the first four layers of ResNet34 with enhanced residual convolutional blocks \cite{mambavision_paper}. These blocks capture fine spatial features more effectively, particularly for edges and fine details, benefiting from pre-trained parameters for faster convergence and improved accuracy.

\noindent\textbf{Experiment 2: Addition of Depth Branch in Dual-Stream Architecture} \\ 
Building on Experiment 1, we introduced a depth branch in a dual-stream setup. With pre-trained parameters from a large-scale model, this branch enhances the model’s geometric understanding of the scene, resulting in better spatial consistency and feature representation.

\noindent\textbf{M3D: Full Feature Enhancement with Dual-Stream and SSM-Enhanced Deep Features} \\
In the final experiment, we replaced ResNet18-based deep feature extraction with a selective attention module to capture long-range spatial dependencies and global consistency. This combination and enhanced shallow features allow for more robust spatial modeling, leading to higher accuracy and fidelity in 3D reconstruction.

\subsection{Ablation Experimental Results}

\noindent\textbf{Learning Rate Impact.} To evaluate the impact of different learning rates on our M3D model, we conducted a learning rate ablation study. We trained the model using two learning rates for this ablation: 0.001 and 0.00006. The learning rate of 0.001 was chosen as a commonly used baseline \cite{SSR} in 3D reconstruction tasks with convolutional networks such as ResNet \cite{resnet}.

\begin{table}[ht]
\centering
\caption{Comparison of Training Loss at Different Learning Rates (lr) Across Epochs in Validation Datasets}
\label{tab:lr_ablation}
\renewcommand{\arraystretch}{1.6} 
\resizebox{\columnwidth}{!}{\small{ 
\begin{tabular}{c|ccccc}
\hline
\rowcolor{LightGray} 
\textbf{lr / epoch} & \textbf{0}\text{\small (base)} & \textbf{5} & \textbf{10} & \textbf{15} & \textbf{20} \\ 
\hline
0.001 & 0.156 & 0.154 \textcolor{Green}{(-1.3\%)} & 0.154 \textcolor{Green}{(-1.3\%)} & 0.151 \textcolor{Green}{(-3.2\%)} & 0.151 \textcolor{Green}{(-3.2\%)} \\
\rowcolor{LightGray} 
\textbf{0.00006} & \textbf{0.176} & \textbf{0.118} \textcolor{Red}{(-32.9\%)} & \textbf{0.100} \textcolor{Red}{(-43.2\%)} & \textbf{0.091} \textcolor{Red}{(-48.3\%)} & \textbf{0.084} \textcolor{Red}{(-52.3\%)} \\
\hline
\end{tabular}
}}
\end{table}

The results in Tables \ref{tab:lr_ablation} demonstrate that using a learning rate of 0.00006 results in approximately \textbf{44.4\% lower} training loss by epoch 20 compared to 0.001, confirming that a lower learning rate enhances convergence and accuracy. Additionally, the learning rate of 0.001 fails to effectively reduce the training loss over epochs, indicating that it is insufficient for achieving stable convergence in our model.

\begin{table}[t]
  \centering
 \caption{Evaluation of CD, F-Score, and NC Across Different Epochs (30 - 120) in Test Datasets. Percentage values next to each metric indicate performance improvement at final evaluation compared to the baseline, corresponding to data in Fig.~\ref{fig:ablation_experiments}.}

  \label{tab:combined_evaluation}
  \renewcommand{\arraystretch}{1.2} 
  \resizebox{\columnwidth}{!}{ 
    \begin{tabular}{@{}lccccc@{}}
      \toprule
      \multicolumn{6}{c}{\textbf{CD $\downarrow$ \textcolor{Red}{(36.9\%)}}} \\
      \midrule
      \textbf{Experiments} & epoch\_30 & epoch\_50 & epoch\_70 & epoch\_100 & epoch\_120 \\
      \midrule
      Baseline (SSR) \cite{SSR} & \underline{23.62} & 19.96 & 17.62 & 15.55 & 16.51 \\
      Exp 1 \textcolor{Red}{(+13.8\%)}    & 27.74 & \underline{16.64} & 18.78 & 16.91 & 14.23 \\
      Exp 2 \textcolor{Red}{(+28.6\%)}    & 27.19 & 19.07 & \underline{14.55} & \underline{12.06} & \underline{11.79} \\
      \textbf{M3D \text{\small (ours)}} \textcolor{Red}{(+53.9\%)}     & \textbf{17.07} & \textbf{11.27} & \textbf{9.77} & \textbf{8.60} & \textbf{7.61} \\
      \midrule
      \midrule 
      \multicolumn{6}{c}{\textbf{F-Score $\uparrow$ \textcolor{Red}{(13.3\%)}}} \\ 
      \midrule
      \textbf{Experiments} & epoch\_30 & epoch\_50 & epoch\_70 & epoch\_100 & epoch\_120 \\
      \midrule
      Baseline (SSR) \cite{SSR} & 0.580 & 0.585 & 0.590 & 0.629 & 0.596 \\
      Exp 1 \textcolor{Red}{(+10.9\%)}    & 0.540 & 0.620 & 0.623 & 0.646 & 0.661 \\
      Exp 2 \textcolor{Red}{(+19.6\%)}    & \underline{0.604} & \underline{0.653} & \underline{0.696} & \underline{0.709} & \underline{0.713} \\
     \textbf{ M3D \text{\small (ours)}} \textcolor{Red}{(+32.7\%)}     & \textbf{0.659} & \textbf{0.743} & \textbf{0.765} & \textbf{0.776} & \textbf{0.791} \\
      \midrule
      \midrule 
      \multicolumn{6}{c}{\textbf{NC $\uparrow$ \textcolor{Red}{(5.5\%)}}} \\ 
      \midrule
      \textbf{Experiments} & epoch\_30 & epoch\_50 & epoch\_70 & epoch\_100 & epoch\_120 \\
      \midrule
      Baseline (SSR) \cite{SSR} & 0.805 & 0.811 & 0.819 & 0.815 & 0.794 \\
      Exp 1 \textcolor{Red}{(+5.7\%)}     & 0.807 & 0.826 & 0.831 & 0.835 & 0.839 \\
      Exp 2 \textcolor{Red}{(+9.3\%)}     & \underline{0.840} & \underline{0.849} & \underline{0.864} & \underline{0.866} & \underline{0.868} \\
      \textbf{M3D \text{\small (ours)}} \textcolor{Red}{(+12.7\%)}     & \textbf{0.852} & \textbf{0.878} & \textbf{0.885} & \textbf{0.891} & \textbf{0.895} \\
      \bottomrule
    \end{tabular}
  }
\end{table}

\noindent\textbf{Training Convergence Analysis in Ablation .}
The primary objective of this experiment is to evaluate the convergence rate of our proposed M3D model compared to other configurations. As shown in Table~\ref{tab:combined_evaluation}, we observe that certain modules (Exp1, Exp2) initially exhibit slower convergence rates. This might be attributed to the limited parameters in the shallow convolutional layers, which take longer to adapt to the 3D reconstruction task. However, these configurations achieve accelerated convergence by the 50th epoch. In contrast, our M3D model, incorporating the state-space model (SSM), shows strong adaptability to the 3D reconstruction requirements from the onset. It maintains a consistently faster convergence rate than the baseline (SSR) from the early stages up to the 120th epoch.

The superior performance of the M3D model is likely due to the SSM's ability to capture both local and global contextual features, which enhances its robustness during training. This capability allows the model to quickly align with the optimal reconstruction direction, resulting in both high-quality reconstructions and faster convergence. Overall, our M3D framework achieves state-of-the-art (SOTA) performance, demonstrating substantial improvements in both convergence speed and 3D reconstruction quality over the baseline.

\begin{figure}[t]
  \centering
  \includegraphics[width=0.9\linewidth]{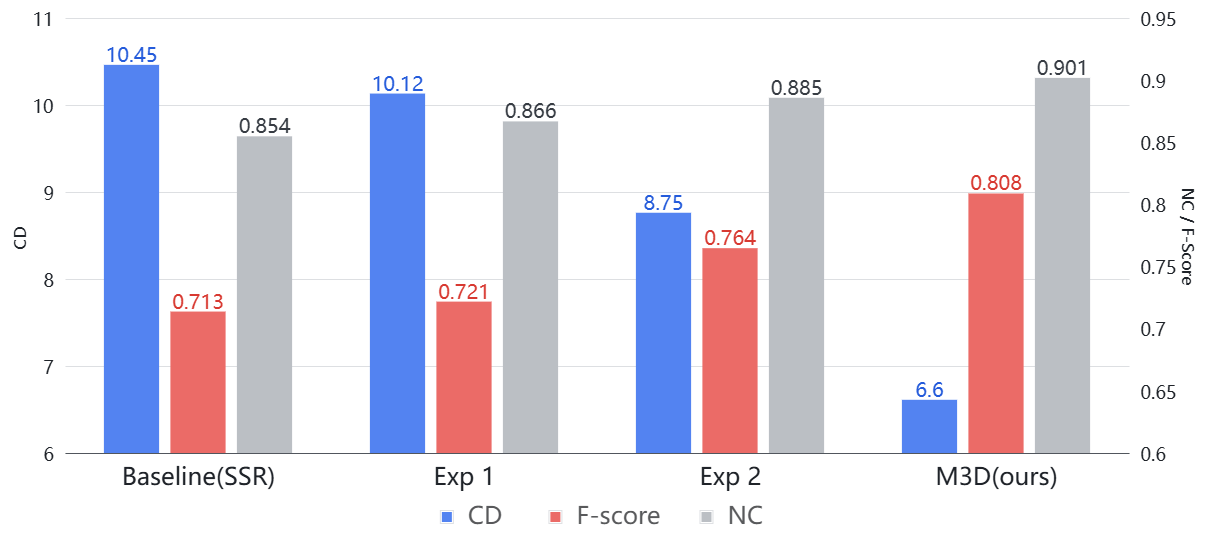}
  \caption{Ablation experiments comparing the performance of different configurations. F-Score values are divided by 100 for visualization purposes to align with the NC axis.}

  \label{fig:ablation_experiments}
\end{figure}

\noindent\textbf{Ablation Evaluation Analysis.} As shown in Figure~\ref{fig:ablation_experiments}, we select SSR as the baseline method for comparison in our ablation study due to its superior metrics and similar reconstruction structure, which primarily relies on ResNet for feature extraction. This choice enables a fair comparison to assess the contributions of each module in our proposed M3D framework. The results demonstrate a progressive improvement in performance with each module added to our design. Starting with the Enhanced Shallow Feature Extraction module, followed by the Dual-Stream structure, and culminating in our full M3D framework, each component consistently contributes positively to 3D reconstruction performance. The steady improvement across all metrics further confirms the positive contribution of each module to reconstruction quality, highlighting the efficiency and reliability of our proposed method. It also demonstrates the coordination among the various modules within our approach.

\section{Conclusion}

In this paper, we proposed M3D, a framework for single-view 3D reconstruction that addresses global and local feature integration limitations. By leveraging a dual-stream architecture that incorporates depth-guided geometric features and selective SSM for enhanced contextual modeling, M3D achieves high-fidelity reconstructions with remarkable accuracy and detail. Our ablation studies validate the contributions of each module, demonstrating substantial improvements in reconstruction metrics with depth information, optimized residual blocks, and advanced feature fusion.  Given the scarcity of comprehensive 3D datasets, we aim to reduce data dependency by exploring semi-supervised learning for 3D reconstruction. This advancement could significantly benefit the field, making high-quality 3D reconstruction more accessible for applications in areas such as virtual reality and autonomous driving.

{
    \small
    \bibliographystyle{ieeenat_fullname}
    \bibliography{main}
}

\end{document}